\documentclass[a4paper, 10 pt, conference]{IEEEtran}
\usepackage{amsmath,graphicx}
 \usepackage{algorithm}
\usepackage{algpseudocode}
\usepackage{footnote}
\usepackage{placeins}
\usepackage{amssymb}
\usepackage{amsmath}
\usepackage{textcomp}
\usepackage{multirow}
\usepackage{placeins}
\usepackage{footnote}
\usepackage{tabulary}
\usepackage[table,xcdraw]{xcolor}
\usepackage{cite}
\usepackage{booktabs}

\DeclareMathOperator*{\argmax}{argmax}
\newcommand*{\argmaxl}{\argmax\limits}

\begin{document}

\title{Object Classification using Ensemble of Local and Deep Features}
%
%\titlerunning{Hamiltonian Mechanics}  % abbreviated title (for running head)
%\author{Kamlesh Jaiswal \and Siddharth Srivastava \and
%Prerana Mukherjee \and Brejesh Lall }
%
%\authorrunning{Kamlesh Jaiswal et al.} % abbreviated author list (for running head)
% \author{Siddharth Srivastava, Prerana Mukherjee, 
%        Brejesh Lall, and Kamlesh Jaiswal}

% \institute{Indian Institute of Technology, Delhi, India.\\
% \email{\{eez127506, eez138300, brejesh}@ee.iitd.ac.in\}, fetkamlesh@gmail.com}

 \author{\IEEEauthorblockN{Siddharth Srivastava\textsuperscript{*},
Prerana Mukherjee\textsuperscript{*},
Brejesh Lall and Kamlesh Jaiswal
 }
\IEEEauthorblockA{Department of of Electrical Engineering, Indian Institute of Technology Delhi \\Email: \{eez127506, eez138300, brejesh\}@ee.iitd.ac.in, fetkamlesh@gmail.com}}

\maketitle              % typeset the title of the contribution

\begin{abstract}
In this paper we propose an ensemble of local and deep features for object classification. We also compare and contrast effectiveness of feature representation capability of various layers of convolutional neural network. We demonstrate with extensive experiments for object classification that the representation capability of features from deep networks can be complemented with information captured from local features. We also find out that features from various deep convolutional networks encode distinctive characteristic information. We establish that, as opposed to conventional practice, intermediate layers of deep networks can augment the classification capabilities of features obtained from fully connected layers.
%\keywords{deep ensemble, SIFT, fisher vector, classification}
\end{abstract}

\section{Introduction}
\footnote{Equal Contribution}
Millions of images are uploaded daily over the Internet. Most of these images are untagged yet have potential to assist in many interesting applications. Manual management of this rapidly increasing data is a humongous task. This large amount of information not only needs to be managed but organized  as  well. Image classification is a problem which aims at organizing the images into various categories based on the information present in them. With the increasing quality of captured images, the classification methods also need to get robust and faster. This in turn supports effective indexing and retrieval of images. There are several applications which are based on object classification: Image search, Video search, robotics etc. Moreover, there have been extensive efforts by the research community to promote image classification problem by hosting competitive datasets (Ex: Pascal, ImageNet) year after year. Since the work of Alex Krizhevsky \cite{krizhevsky2012imagenet} in 2012 which won the ImageNet challenge, CNNs are the state of the art in this domain. Therefore, it becomes imperative to not only explore it further but also attempt to extract the strengths of various feature combinations, local or global, and utilize them for effective classification. Despite excellent performance of the state of the art techniques on various datasets, we are still far from attaining human level performance as 2D images have limited information and context is as important as content.

Although Convolutional Neural Networks have become very popular for computer vision tasks such as digit recognition \cite{niu2012novel}, object recognition \cite{redmon2016you}, object classification \cite{redmon2016you} etc., where the reason for success in such tasks can be primarily attributed to their ability to learn relevant features from the input data as against the traditional approaches of hand-crafted features. In general, a CNN is augmented with a softmax layer for performing classification. More recently, many state of the art results have been obtained using CNN as a feature extraction technique and Support Vector Machines (SVM) as a classifier. Various works claim the long proven capability of SVM for classification tasks and its comparatively better generalization ability as a reason for such augmentation. 

While using output from the last layer of various CNN architectures has become a standard feature for many computer vision tasks, some studies \cite{kataoka2015feature} suggest that intermediate layers of CNN may also be suitable or in some cases better than the output of the last layer. In this context, we evaluate capability of features from intermediate layers of various popular CNN architectures while also evaluating the strength by concatenating features from multiple layers. Moreover, Yang et. al. \cite{lenc2015understanding} demonstrated that CNN features are not invariant to various image level transformations. Therefore, we compare feature matching capability of CNN with SIFT. Gaining insights from these experiments, we propose an ensemble of features consisting of deep features from various CNN architectures with SIFT. We argue that while deep features provide robustness, the local features augment the capability of deep features by providing robustness and consistency against transformations of objects in complex scenes. The argument also obtains support from works of \cite{nguyen2015deep}, which show that the features from CNN characterize objects in terms of relative organization of patterns of colors etc. whereas local features provide characterization of a patch thus supplementing for such pitfalls.

In view of the above, the contributions of this work are: (i) We present an end-to-end framework for image classification using features from Convolutional Neural Network (ii) We comprehensively evaluate the feature representation strength of intermediate layers of various CNN architectures providing many useful insights.

\section{Related Work} 
Razavian et. al. \cite{sharif2014cnn}, through rigorous experiments suggest that deep convolutional features should be primary features for vision related tasks. Conventionally, the scientific works utilizing deep convolutional features consists of output of the last fully connected layer from a single network or a cascade of networks. Most of the techniques utilizing multiple convolutional neural networks require retraining the network. However, an alternate approach which has been especially successful with handcrafted features is to fuse features \cite{srivastava2016characterizing} or form ensembles to collectively harness the strength of multiple features during classification. Along the same lines, in this research, we work with deep features to form ensemble of classifiers. Ensemble of multiple convolutional neural network models has been explored earlier in literature such as by Krizhevsky et. al. \cite{krizhevsky2012imagenet}, Zeiler and Fergus \cite{zeiler2014visualizing}, Simonyan et. al. \cite{simonyan2014two}. Such ensembles have resulted in improved classification rate primarily due to complementarity of various models with modifications in the same network. In this work, we exploit the complementarity of features from distinct networks instead of differently trained models of the same network. Majtner et. al \cite{majtner2016combining} train two SVM classifiers for skin lesion classification. First classifier is trained with concatenation of Surf and LBP features while the second classifier is trained with features from AlexNet. The final decision is assigned as the output based on the classifier with higher classification confidence score. Although, this work is closest to ours in terms of exploiting the combination of deep and local features, the proposed work differs in the following aspects: (i) Firstly, instead of fusing features, we train individual classifiers for each feature (ii) Instead of assigning the output as the highest confidence score, we allow each classifier to participate in the decision process by voting up for the final decision.

Authors in \cite{cohen2014transformation, lenc2015understanding} show that the output of the convolution layers are not invariant to large image transformations. It is easy to extend the same inference to the output of the fully connected layers since they feed on to the output of the intermediate layers. Jaderberg et. al. \cite{jaderberg2015spatial} alleviate this problem with Spatial Transformer Network which can be added to existing CNN architecture. However, such models need to be trained along with modifications to the existing architectures. In this work, we leverage SIFT features \cite{lowe1999object}, which are designed to be invariant to image transformations. Perronnin et. al. \cite{perronnin2015fisher} show that using encoded local features with fully connected layers is computationally less expensive than CNN while outperforming traditional approaches. In this work we encode SIFT using Fisher Vectors \cite{sanchez2013image} and predict the output with a majority voting scheme over an ensemble of SVM classifiers trained on the discussed features. The Fisher Vector with SIFT provides robustness by voting up with high confidence for images which may be misclassified by deep features due to transformational artifacts. With this premise, we discuss various architectures and the methodology in the next section.

\section{Methodology}

\begin{figure*}[thpb]
      \centering
      \fbox{
      \includegraphics[scale=.5]{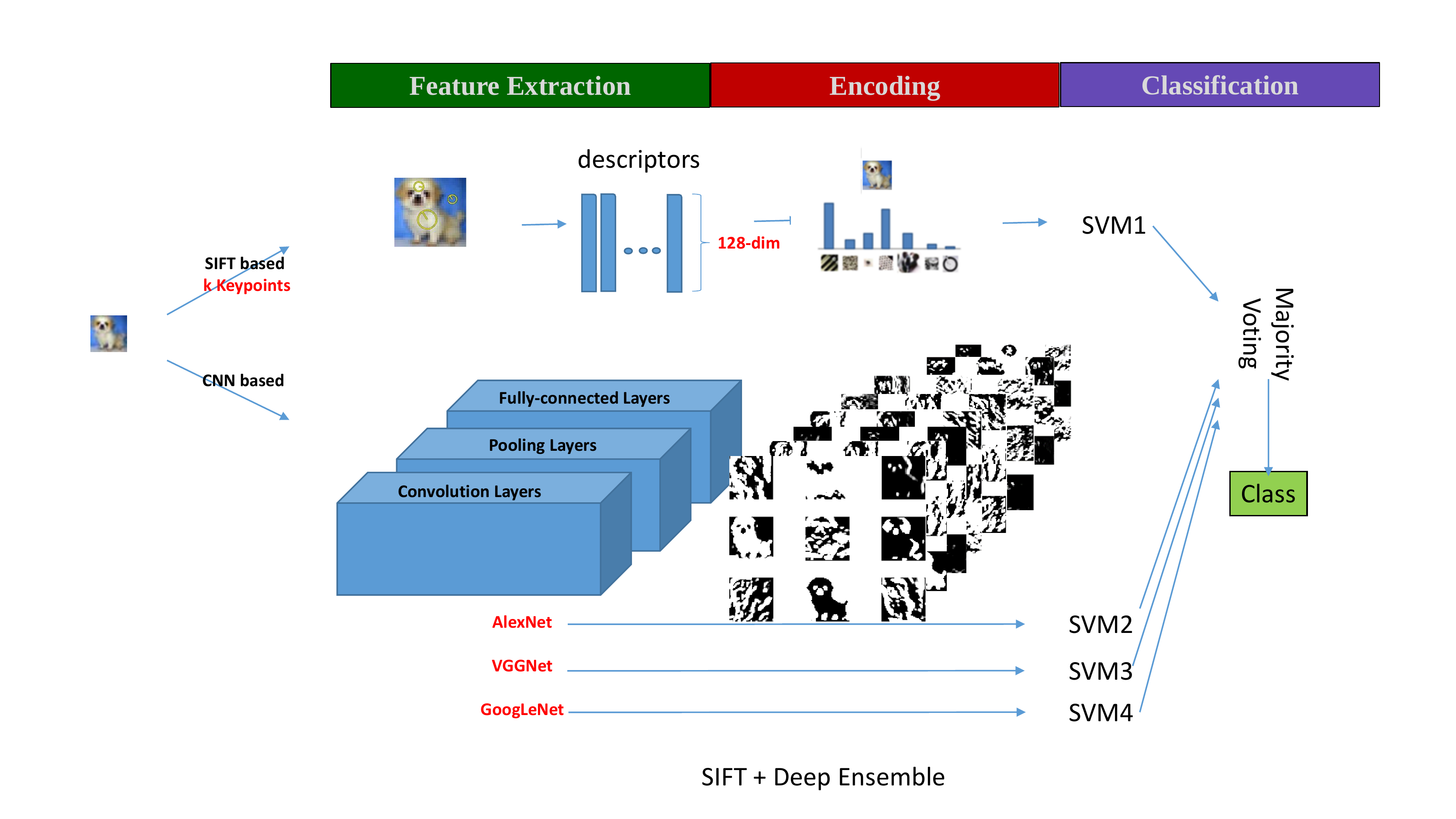}}
      \caption{Architecture of ensemble of features}
      \label{fig:workflow}
\end{figure*}

The overall architecture is shown in Figure \ref{fig:workflow}. We extract descriptors from last fully connected layer of deep networks along with Fisher Vector of SIFT descriptors. For each feature, we train an SVM optimized for classification on the input dataset followed by a majority voting scheme to arrive at the final decision. We now describe each component in detail.  

\subsection{Deep Convolutional Features}

In this work, we evaluate three popular CNN architectures: AlexNet \cite{krizhevsky2012imagenet}, VGGNet  \cite{Simonyan14c} and GoogleNet \cite{szegedy2015going}.

\subsubsection{AlexNet}
The major leap in the advancements of deep architecture came with the advent of AlexNet architecture proposed by Krizhevsky et. al \cite{krizhevsky2012imagenet} in 2012. It completely changed the perspective of neural networks and went to win the large scale ImageNet challenge. The reason for its unparalleled performance was to employ of NVIDIA GPU to reduce the training time (providing a speedup of 10x) and scaling to deep architecture led to the learning of intricacies of the object hierarchies and understanding complex scene environment. 

The architecture consists of $5$ convolution layers, $3$ fully connected layers. It introduced the use of rectified linear units (ReLU) as non-linearities in the pooling layer and dropout to ignore neurons during training, thus reducing overfitting. The pooling layers are placed after first, second and fifth convolution layers. 

\subsubsection{VGGNet}
The popularity of VGGNet \cite{Simonyan14c} is primarily due to the use multiple $3$x$3$ filters in each convolutional layer. These multiple small convolutional filters can mimic the response of a large receptive field thus providing better generalization capability and represent complex features of the objects. 

VGGNet  consists of 16 layers with $13$ convolution layers and $3$ fully connected layers.  The convolution layers are divided into groups of $5$ with each group followed by a max pooling layer. VGGNet uses huge parameters in different layers and thus inference is quite costly at run-time. This was handled by the Inception model which was computationally less expensive and was introduced in GoogleNet as described next.

\subsubsection{GoogleNet}
GoogleNet introduced the inception metalayer to compensate the overhead of deep convolutional architectures. It parallelizes the convolutional blocks with $1$x$1$ convolutional filters, known as Network-in-network (NiN) blocks. It effectively utilizes very few parameters which are shared across all pixels of these convolutional features. 

\subsection{Local Features and Encoding}
The patch-based image descriptors like SIFT, SURF show huge potential in image classification systems due to the capability of these features to contain more information about the image content than the raw pixels and the invariance properties achieved. In the keypoint detection stage, an appropriate scale for feature is chosen as the continuous function of scale $\sigma$ to form a scale space of the image by convolving it with a Gaussian kernel. The best notion of the scale is determined by the maxima of a Laplacian-of-Gaussian filter. This can be replaced by a Difference of Gaussian operator. The extrema points of the keypoints are estimated using a neighborhood operator of $3$x$3$ filter. A further filtering stage helps in getting rid of non-true extrema points, low contrast points and along edge responses. In order to make the descriptor rotation invariant an orientation histogram is computed and the final keypoint is described with a 128- dimensional feature vector. Several variants of SIFT like Color SIFT, Root Sift, PCA-SIFT were also introduced but they did not gain the same popularity as compared to SIFT. Deep features have been the most sought after features in the current era of computer vision. We perform a combination of these features and analyze the performance boost obtained by the complementary attributes provided by both these set of features in Section \ref{sec:results}. 

The fine-grained information of the images is captured using image signatures (bag of visual words, VLAD, fisher vector). In our experiments, we utilize fisher vector as the encoding strategy. Given a likelihood function p($X|\lambda$)where $\lambda$ denotes the parameters, the score function of a sample $X$ can be given as:
\begin{equation}
G_\lambda^X=\triangledown_{lambda} log p(X|\lambda) 
\end{equation}
The gradient vector can be classified using any discriminative classifier. It is required to normalize the inner product term present in such discriminative classifiers. The Fisher information matrix is used for this purpose is given by,
\begin{equation}
F_{lambda}=E_X[\triangledown_{lambda} log p(X|\lambda) \triangledown_{lambda} log p(X|\lambda)']
\end{equation}
The normalized gradient vector is thus given by
\begin{equation}
F_{lambda}^{-1/2} \triangledown_{lambda} log p(X|\lambda)
\end{equation}
Fisher kernels on visual vocabularies are represented using Gaussian Mixture Models (GMM). 

\subsection{Ensemble of Classifiers}

Here we explain various architectures proposed and evaluated in this work. We split the explanation into training and testing phases. We begin by explaining the feature construction and training methodology for each architecture followed by formulation of test phase. 

\subsubsection{Training}
\begin{itemize}
\item \textbf{Deep Ensemble}: The deep networks are trained using the respective softmax classifier at the last layer. However, Tang et. al \cite{tang2013deep} demonstrate that using SVM instead of softmax yields better results. Since we are interested in utilizing the feature representation capability of various deep networks, therefore we replace softmax with SVM in the last layer and retrain the last layer with the output of the fully connected layers. We term the combination of various deep features, with independently trained SVMs as \textit{Deep Ensemble}. Such a network allows exploiting the complementarity of various deep features.

\item \textbf{Ensemble of Intermediate Layers}: 
\begin{itemize}
\item \textit{Individual intermediate layers}: For evaluating the representation capability of the intermediate layers, we remove the subsequent layers once the network has been trained. Thereafter, the respective intermediate layer is followed by the classification layer consisting of SVM.
\item \textit{Fusion of intermediate layers}: For each deep network, we also perform experiments on fusion of features from various intermediate layers as shown in Fig \ref{fig:workflow1}. This is based on the intuition that intermediate layers capture various levels of information about an image. Such a combination is evaluated to compare and gain insights if the fusion of various intermediate layers can form a stronger feature as compared to \textit{Deep Ensemble}. Since the resulting feature vector has very high dimensionality, we train the SVM by reducing the size of the feature vector using PCA.
\end{itemize}

\begin{figure*}[thpb]
      \centering
      \fbox{
      \includegraphics[scale=.5]{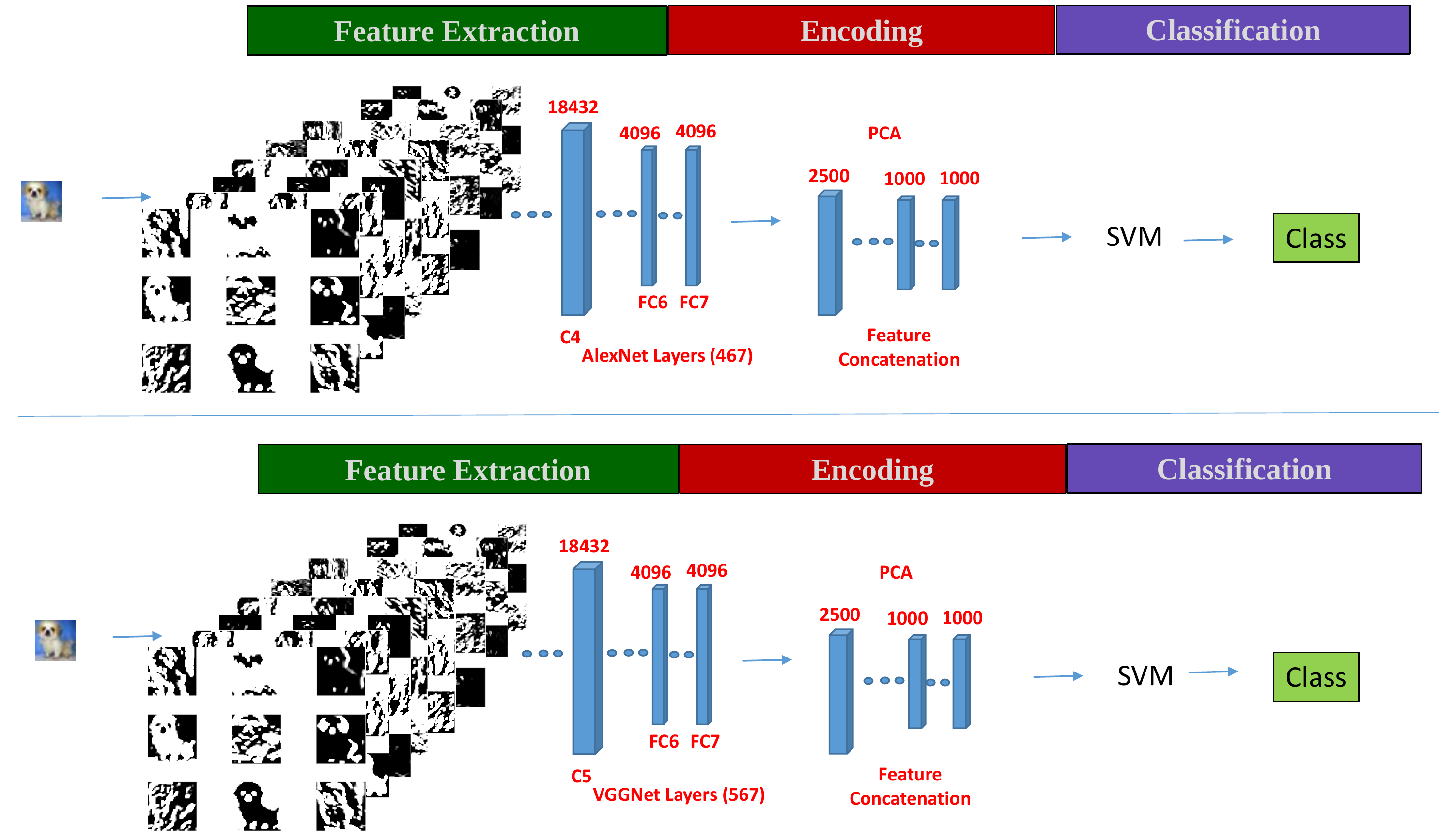}}
      \caption{Ensemble of intermediate layers}
      \label{fig:workflow1}
\end{figure*}

\item \textbf{SIFT with Deep Ensemble}: The third architecture consists of fusing the output of SIFT with Deep Ensemble. We quantize the SIFT features from the images using Fisher Vector. The Fisher Vectors are obtained on the training set using the methodology of Perronnin et. al \cite{perronnin2010improving}. We combine the output of the SVM thus obtained with \textit{Deep Ensemble} as discussed in the next section. 
 
\end{itemize}

\subsubsection{Testing}
At test time, the output class for various architectures discussed previously is predicted based on majority voting performed as follows.

\begin{equation}\label{eq:majvoting}
\phi_{ens}(I) = \argmaxl_{k} N_{k}
\end{equation}

where $\phi_{ens}(I)$ is the output decision for an input image $I$, $N_{k}$ is the number of SVMs whose output is the $k^{th}$ class and is given by

\begin{equation}
N_{k} = \#\{c | \phi_{c}(I) = L_k \}
\end{equation}

Here, $\phi_{c}$ is the output or decision function for $c^{th}$ classifier in the ensemble and $L_{k}$ denotes the label for $k^{th}$ class $\forall k \in [1,K]$.

\section{Experiments and Results}\label{sec:results}

\subsection{Experimental Setup}
The experiments were performed on a system with i5 processor with 32GB of RAM and 1GB Nvidia Quadro Graphic Card. The implementation was done in MatConvNet \cite{vedaldi2015matconvnet}. We used pretrained models for AlexNet, VGGNet and GoogleNet from MatConvNet repository. We used the SIFT implementation of VLFeat \cite{vedaldi2010vlfeat}. The SVM implementation was used from libSVM \cite{chang2011libsvm}.

\subsection{Dataset}
We provide results on CIFAR-10 \cite{krizhevsky2009learning}. It consists of objects from $10$ classes with $60,000$ images in total. We use the standard train and test split of the dataset for reporting results in this paper. 

\subsection{Results and Discussion}

\textit{Baseline}: In order to evaluate the performance gains obtained using the proposed architectures, we perform experiments on vanilla deep architectures along with various combinations. 
As discussed, we also performed experiments using features from intermediate layers of various CNN architectures. The list of CNN architectures, layers and corresponding dimensionality prior and post applying PCA are shown in Table \ref{tab:dimensions}.

\textit{Observations}: In Table \ref{tab:svm} we compare the proposed ensemble approach against various combination of features from different CNN architectures. It can be observed that VGGNet (6) performs better than other VGGNet features. Since, VGGNet (6) represents the penultimate layer of the architecture, it indicates that the last fully connected layer results in loss of feature distinctiveness. However, the same is observed for AlexNet (4)  with both raw and PCA reduced descriptors where the accuracy is highest among the considered layers of AlexNet being $87.1\%$ and $88.3\%$ respectively. Moreover, higher accuracy with PCA for AlexNet (4) demonstrates that $4^{th}$ layer, which is the last convolution layer has redundant features and further layers reduce the strength of the descriptor. But it would be important to note that while $4^{th}$ layer provides highest accuracy, the size of the raw descriptor is nearly $4$ times the subsequent layers while we still achieve a $3.5\%$ higher mean accuracy than other PCA reduced AlexNet descriptors. 

VGGNet (567) and AlexNet (457) show the results after PCA driven reduction. The results of raw feature concatenation on SVM is not provided as the dimensionality of the features become too large. It can be seen that such a combination on an average achieves approximately $3\%$ improvement over other AlexNet and VGGNet features which is a significant gain given that no additional complexity has been introduced for combining or fine-tuning the descriptors. The table also shows the results using the proposed Deep Ensemble and (SIFT + Deep Ensemble) approaches. The former shows an average improvement of $4.5\%$, $4.2\%$ and $8.8\%$ over $7^{th}$, $6^{th}$ and $5^{th}$/$4^{th}$ layers of vanilla VGGNet and AlexNet architectures respectively. Similarly, the (SIFT+ Deep Ensemble) results in respective improvements of $4.8\%$, $4.5\%$, $9.2\%$. 

\begin{table}[]
\centering
\begin{tabular}{@{}lcc@{}}
\toprule
\textbf{CNN Model (Layer)} & \multicolumn{1}{l}{\textbf{Dimension}} & \multicolumn{1}{l}{\textbf{Dimension (PCA)}} \\ \midrule
AlexNet (4)                & 18432                                  & 2500                                                 \\
AlexNet (5)                & 4096                                   & 1000                                                 \\
AlexNet (7)                & 4096                                   & 1000                                                 \\
VGGNet (5)                 & 18432                                  & 2500                                                 \\
VGGNet (6)                 & 4096                                   & 1000                                                 \\
VGGNet (7)                 & 4096                                   & 1000                                                 \\ \bottomrule
\end{tabular}
\caption{CNN Model and respective feature dimensions. Second column lists the actual dimension of the feature while the third column lists the length of feature obtained after applying PCA}
\label{tab:dimensions}
\end{table}

% \usepackage{booktabs}
\begin{table}[]
\centering
\begin{tabular}{@{}lcc@{}}
\toprule
\textbf{CNN Model (Layer)} & \multicolumn{1}{l}{\textbf{Accuracy (SVM)}} & \multicolumn{1}{l}{\textbf{Accuracy (PCA+SVM)}} \\ \midrule
VGGNet (7)                 & 87.6                                        & 86.9                                            \\
VGGNet (6)                 & 90.1                                        & 88.3                                            \\
VGGNet (5)                 & 80.1                                        & 85.9                                            \\
AlexNet (7)                & 86.1                                        & 86.5                                            \\
AlexNet (6)                & 84.3                                        & 84.2                                            \\
AlexNet (4)                & 87.1                                        & 88.3                                            \\
VGGNet (567)               & -                                           & 89.8                                            \\
AlexNet (457)              & -                                           & 88.9                                            \\
Deep Ensemble              & 90.8                                           & -                                            \\
SIFT + Deep Ensemble              & \textbf{91.1}                                           & -                                            \\
\bottomrule
\end{tabular}
\caption{Classification Accuracy (\%) of various CNN models. VGGNet (567) represents the concatenation of features from layers $5^{th}$,$6^{th}$ and $7^{th}$ while AlexNet (467) represents the concatenation of features from $4^{th}$, $5^{th}$ and $7^{th}$ layers}
\label{tab:svm}
\end{table}

\section{Conclusion}
We proposed and evaluated an ensemble of local and deep features for object classification. Additionally, we compared feature representation capability of various intermediate layers of convolutional neural network. We performed extensive evaluation on CIFAR-10 dataset and demonstrated that local features such as SIFT can complement the deep features. We also found that different deep architectures characterize distinctive information of an image. Additionally, we evaluated features from intermediate layers and there combination, which led us to conclude that such features also complement features from fully connected layers. 

%\clearpage
\bibliographystyle{IEEEtran}    
\bibliography{biblio}

\end{document}